\pdfoutput=1
\documentclass[final]{nesy2026}

\usepackage[T1]{fontenc}
\usepackage[utf8]{inputenc}
\usepackage{microtype}

\usepackage{listings}
\usepackage{xcolor}
\usepackage{spverbatim}
\usepackage{multirow}
\usepackage{booktabs}

\title[Program Semantic Inequivalence Game]{Program Semantic Inequivalence Game with Large Language Models}

\clearauthor{\Name{Antonio Valerio {Miceli Barone}} \Email{amiceli@ed.ac.uk}\\
  \Name{Vaishak Belle} \Email{vbelle@ed.ac.uk}\\
  \addr University of Edinburgh, Edinburgh, United Kingdom
  \AND
  \Name{Ali Payani} \Email{apayani@cisco.com}\\
  \addr Cisco Systems, San Jose, CA, United States}

\begin{document}
\maketitle
\begin{abstract}
Large Language Models (LLMs) can achieve strong performance on everyday coding tasks, but they can fail on complex tasks that require non-trivial reasoning about program semantics.
Finding training examples to teach LLMs to solve these tasks can be challenging.

In this work, we explore a method to synthetically generate code reasoning training data based on a \textbf{semantic inequivalence game} (\textbf{SInQ}): a \textbf{generator} agent creates program variants that are semantically distinct, derived from a dataset of real-world programming tasks, while an \textbf{evaluator} agent has to identify input examples for which they behave differently. The agents train each other semi-adversarially, improving their ability to understand the underlying logic of code.

We evaluated our approach on multiple code generation and understanding benchmarks, including cross-language \textbf{vulnerability detection} \citep{CodeXGlue2021}, where our method improves vulnerability detection in C/C++ code despite being trained exclusively on Python code, and the challenging \textbf{Python builtin identifier swap} benchmark \citep{miceli-barone-etal-2023-larger}, where modern LLMs still struggle and our approach yields substantial improvements for one of the two base models we fine-tune.

We release the code needed to replicate the experiments, as well as the generated synthetic data, which can be used to fine-tune LLMs\footnote{\url{https://github.com/Avmb/semantic_neq_game}}.
\end{abstract}

\section{Introduction}
\label{SEC:INTRO}

\begin{figure}[ht]
\begin{center}
\includegraphics[width=0.45\linewidth]{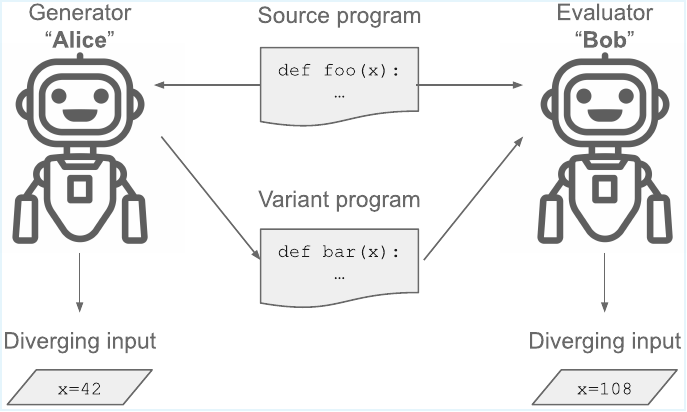}
\end{center}
\caption{Semantic inequivalence game: Alice receives a source program $P$ and generates a variant program $Q$ and a diverging input. Bob receives $P$ and $Q$ and generates another diverging input.}
\label{FIG:METHOD:DIAGRAM}
\end{figure}

Large Language Models (LLMs) are widely used by programmers, but while they perform well on common coding tasks they still struggle with non-trivial reasoning about program semantics \citep{miceli-barone-etal-2023-larger, maveli2025largelanguagemodelscapture}.
This can lead to subtle bugs and to missed vulnerabilities and adversarial backdoors \citep{Dinh2023, dou2024whatswrongcodegenerated}, compromising the safety and security of generated code \citep{wang2024aigeneratedcodereallysafe, mohsin2024trustlargelanguagemodels}.

LLMs' coding capabilities are typically improved by fine-tuning on human-annotated and synthetically generated data. Human annotation is expensive and covers few non-trivial scenarios, while typical synthetic pipelines generate problem statements, solutions and unit tests and validate them by execution: this scales easily but gives limited coverage of problem types and introduces noise, as unit tests often misalign with the problem statements in edge cases.

Self-play trains agents by pitting them against each other, incentivizing them to discover and defend against unusual scenarios, and has reached human-level or superhuman play in Go \citep{silver2016mastering, silver2017alphagozero}, Chess \citep{silver2017alphazero}, Dota 2 \citep{openai2019dota2largescale}, StarCraft II \citep{alphastar2019} and dialogue games such as Diplomacy \citep{metacicero2022}. However, it typically needs external engines to enforce rules and score play, which is hard for open-ended tasks like coding; recreational coding environments such as CROBOTS\footnote{\url{https://github.com/tpoindex/crobots}} are too domain-specific for general code reasoning. We are aware of only one concurrent work, \citet{zhao2025absolutezeroreinforcedselfplay}, that uses self-play to train LLMs for arbitrary code generation, while \citet{dong2025stp} apply a similar idea to theorem proving.

In this work, we introduce a game based on program semantic inequivalence designed to train agents in code reasoning across arbitrary domains.\footnote{An earlier and shorter version of this work \citep{micelibarone2025programsemanticinequivalencegame} was presented at the COLM 2025 AIA workshop.}

By design, this game has no theoretical performance cap.  
We use it to train LLMs through self-play, demonstrating significant performance improvements on challenging tasks.


\section{Proposed Method}
\label{SEC:METHOD}

Our approach pits two LLM agents against each other: the \textbf{generator}, ``Alice'', creates challenging code-understanding problems that the \textbf{evaluator}, ``Bob'', must solve (Figure~\ref{FIG:METHOD:DIAGRAM}). Alice is incentivized to deceive Bob into mistakes, but must also provide solutions, so its instances cannot be unsolvable; training both agents thus drives them towards a deeper understanding of program semantics. The game is based on the \textbf{semantic (in)equivalence} of programs, which allows precise verification of solutions, unlike natural-language specifications or unit tests, and is fundamentally linked to computability theory through reductions to Rice's theorem and the Halting problem. Reasoning about program (in)equivalence is also of practical value for identifying bugs and security vulnerabilities introduced during code refactoring.


\subsection{The semantic inequivalence game}
\label{SEC:METHOD:INEQ}

Two programs \(P\) and \(Q\) are semantically equivalent if, for every input \(x\), they either both halt with the same output or both fail to halt. Determining equivalence is fundamental to program verification and compiler design, but proving it automatically is hard: popular languages such as Java and Python are defined by natural-language specifications or reference implementations rather than formal semantics, and even where formal semantics exist, machine-checkable equivalence proofs for non-trivial code are challenging even for experts. We sidestep this by defining a program-understanding game that focuses solely on \emph{inequivalent} programs.

We define the \textbf{semantic inequivalence game} as the following one-shot interaction between two players: the \textbf{generator}, ``Alice'', and the \textbf{evaluator}, ``Bob'':
\begin{enumerate}
    \item Alice receives a program \(P\) and generates another program \(Q\), which has to be inequivalent to \(P\), along with a diverging input \(x\) such that \(P(x) \neq Q(x)\).  
    \item The diverging input is verified by executing both programs on it. If \(P(x) = Q(x)\), Alice loses.  
    \item Bob receives \(P\) and \(Q\) and attempts to produce a diverging input \(\hat{x}\) (which may or may not be the same as \(x\)). If Bob correctly identifies a diverging input, he wins and Alice loses; otherwise, Bob loses and Alice wins.  
\end{enumerate}  

Correctness is verified simply by executing the programs on the diverging inputs, with no need to generate or check formal proofs. Both agents are trained iteratively through repeated play. The source programs \(P\) are sampled from a dataset of short, self-contained exercises spanning many tasks (e.g.\ MBPP \citep{austin2021programsynthesislargelanguage}), which keeps the game \textbf{grounded} in real-world coding problems; sampling \(P\) rather than letting Alice generate it too removes any incentive for it to produce obfuscated code that would not improve Bob's general reasoning.

To approximate non-termination detection, we impose a randomized time limit that significantly exceeds the typical runtime of source programs. Randomizing it prevents Alice from exploiting a fixed limit, e.g.\ by generating a \(Q\) that simply loops for a predetermined duration before returning \(P\)'s output.


A worked (artificial) example of a single round of the game is given in Appendix \ref{SEC:APPENDIX:EXAMPLE}; Appendix \ref{SEC:APPENDIX:QUALEX} analyses a real example from MBPP.

Unlike Go or Chess, where perfect play is possible, the game has no strict performance cap: under an infinite time limit Bob's task is undecidable (Appendix \ref{SEC:APPENDIX:NDSINEQ}), so in principle the agents can learn arbitrarily complex coding logic while staying grounded in real-world programs. The game is essentially zero-sum when Alice produces only valid outputs, but making it positive-sum can help integration with supervised fine-tuning (SFT) and avoid degenerate strategies (e.g.\ Alice generating cryptographically hard puzzles), as discussed in Appendix \ref{SEC:APPENDIX:TARGETDIFFICULTY}.

\subsection{Implementation with Supervised Fine-tuning and Difficulty Targeting}
\label{SEC:METHOD:DIFFICULTY}

Although the game is well-suited to reinforcement learning, RL was not available on the OpenAI API at the time of our experiments, so we devised the following rejection-sampling fine-tuning implementation with explicit difficulty supervision.

When we present the program pair \((P, Q)\) to Bob, we can sample \(N\) evaluation responses and define the difficulty of the pair based on the number of correct assessments:
\begin{equation*}
    d(P, Q, Bob) = 10 \left( 1 - \frac{N_\text{correct}}{N} \right)
\end{equation*}
For example, if Bob can solve the pair \((P, Q)\) 40\% of the time, the difficulty of this instance is \(6\).

During generation we ask Alice for a program with a specific target difficulty \(d_t\), usually the maximum value of \(10\) (a lower value makes the game positive-sum; see Appendix \ref{SEC:APPENDIX:TARGETDIFFICULTY}). Concretely, we form the prompt \(I = Template_{Alice}(P, d_t)\), sample \(O = Alice(I)\), and extract \((CoT, Q, x) = Extractor_{Alice}(O)\). If the output is invalid we discard it; otherwise we evaluate it with Bob to estimate its actual difficulty and form a training example in which the input's target difficulty is replaced by this estimate (rounded to the nearest integer):
\begin{equation*}
    TE_{Alice} := (Template_{Alice}(P, d(P, Q, Bob)), O)
\end{equation*}
We generate one or more such examples from each source program \(P\) and batch-train Alice (e.g.\ via OpenAI's chat fine-tuning API, with the loss taken on the assistant message \(O\) only).

This dataset can be extended across multiple generations of Alice as long as Bob is unchanged; once Bob is updated (via rejection-sampling SFT on its own successful pairs), the difficulty of every instance must be recomputed against the stronger Bob. Since Alice initially generates examples that are too easy for Bob, especially when both share a base LLM, we found it beneficial to train Alice for many rounds, ideally to convergence, before training each new Bob.

Because Alice's early examples are mostly difficulty-zero, using all of them would swamp the training set with unhelpful instances and minimize the loss on programs we do not want Alice to generate. We therefore bias the set towards \emph{hard} examples (\(d(P, Q, Bob) \geq 5\), i.e.\ fooling Bob at least half the time), adding only a fraction of the rest (20\% of the hard count), sampled without replacement round-robin across difficulty bins.

We also explicitly train Alice to predict instance difficulty, by adding training examples in which Alice must output the difficulty of an instance it has generated; this part of the dataset is biased towards hard examples but retains 50\% easy ones, since we minimize the loss only on the predicted difficulty. We give the exact format of these examples in Appendix \ref{SEC:APPENDIX:DIFFPRED}.

We summarize the overall training procedure in Algorithm \ref{ALG:SINQ} (Appendix \ref{SEC:APPENDIX:ALGORITHM}).

Refer to Appendix \ref{SEC:APPENDIX:TEMPLATES} for all the templates used to interact with the LLM.

\section{Experiments}
\label{SEC:EXPERIMENTS}

\subsection{Training}
\label{SEC:EXPERIMENTS:TRAINING}

We run our main set of experiments using OpenAI \texttt{gpt-4o-mini-2024-07-18} as the base LLM for both Alice and Bob.  
In order to train our agents, we use the training portion of MBPP as our source set of programs, using only the \texttt{code} field from each source example.
We perform additional experiments using OpenAI \texttt{gpt-4.1-nano-2025-04-14} as the base LLM.

We take several precautions against data contamination. As seeds we use only the \texttt{code} field of the MBPP \emph{training} split, never its problem statements or unit tests, and hold out the test split for evaluation. More importantly, seeds only ground the generation of \emph{new}, semantically distinct variants, so memorising MBPP solutions does not by itself help an agent play the game on the resulting pairs. Finally, our headline extrinsic benchmarks are disjoint from MBPP and, for vulnerability detection, in a different language (C/C++), so those gains cannot stem from MBPP memorisation.

We sample \(N = 10\) responses per query at temperature \(1.0\) and \(\text{top\_p} = 0.7\), and validate, normalize and execute the generated programs in a sandbox under a randomized time limit; the generation and validation pipeline is detailed in Appendix \ref{SEC:APPENDIX:EXPDETAILS}.

We train the models via SFT with difficulty targeting (always set to 10) and difficulty prediction, as described in Section \ref{SEC:METHOD:DIFFICULTY}, using the OpenAI fine-tuning platform.
We rely on the platform's default hyperparameters (selected heuristically from the training-set size and base model): 3 epochs in all cases, with batch size 4--7 and learning-rate multiplier \(1.8\) for \texttt{gpt-4o-mini}, and batch size 1--3 and multiplier \(0.1\) for \texttt{gpt-4.1-nano}. Rather than early-stopping on a validation set, we add Alice rounds until the mean instance difficulty against the fixed untrained Bob converges, then run a single Bob round.

\begin{figure}[ht]
\begin{center}
\includegraphics[width=0.43\linewidth]{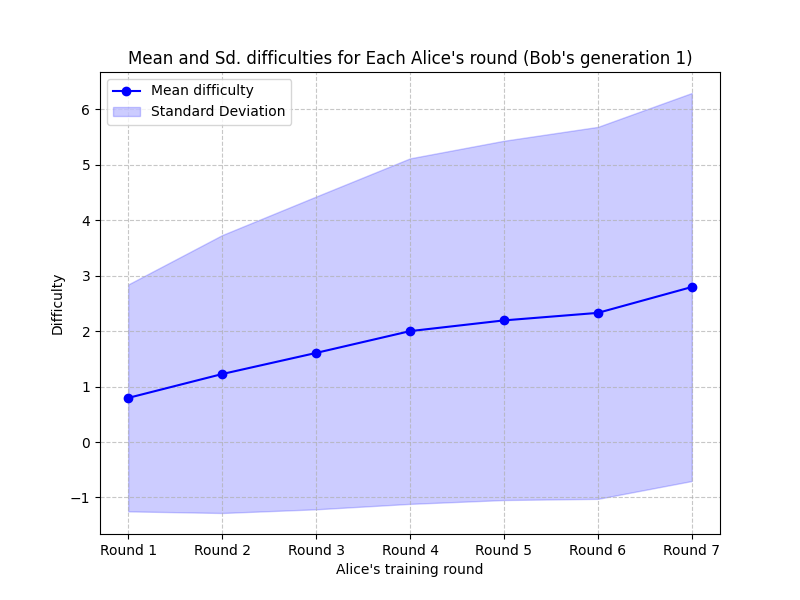}
\includegraphics[width=0.43\linewidth]{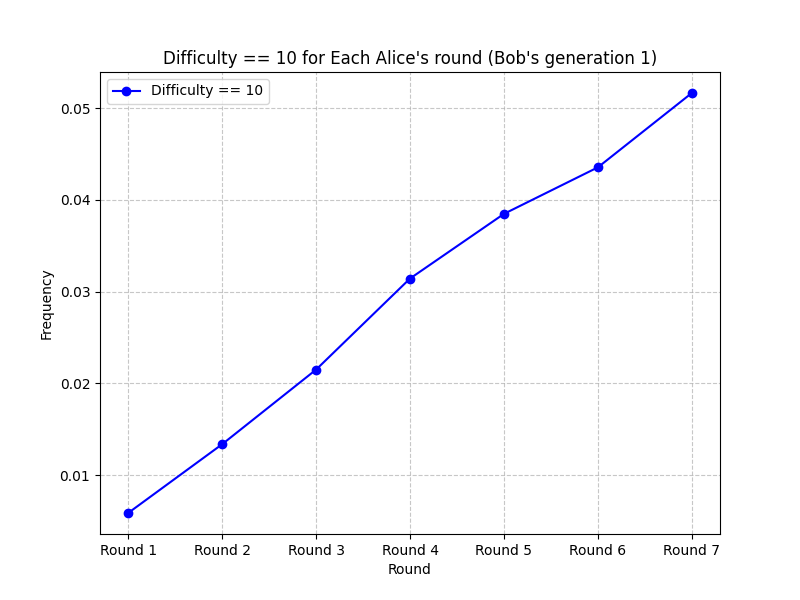}
\end{center}
\caption{Instance difficulty with respect to an untrained Bob, plotted against the number of Alice's training rounds, \texttt{gpt-4o-mini} models. Left: mean and standard deviation. Right: fraction of instances with maximum difficulty.}
\label{FIG:EXPERIMENTS:TRAIN:GPT4_MINI_DIFF_CURVES}
\end{figure}

On \texttt{gpt-4o-mini} we run 7 Alice rounds then one Bob round (fewer than ideal, due to our budget limits), and on \texttt{gpt-4.1-nano} 6 Alice rounds, which suffice for convergence, then one Bob round. Difficulty curves are in Figure \ref{FIG:EXPERIMENTS:TRAIN:GPT4_MINI_DIFF_CURVES} (\texttt{gpt-4o-mini}) and Figure \ref{FIG:EXPERIMENTS:TRAIN:GPT_4_1_NANO_DIFF_CURVES} (\texttt{gpt-4.1-nano}, Appendix \ref{SEC:APPENDIX:NANOCURVES}).

Each Alice round starts from the base LLM checkpoint rather than the previous fine-tune, while accumulating training instances across rounds. Retraining from base avoids over-representing the abundant easy instances from early rounds and prevents distribution drift from compounding across fine-tunes, which matters because the per-round sets are small and re-used. Difficulty estimates stay valid as long as Bob is unchanged; after a Bob update they must be recomputed (or the set discarded) before further Alice rounds.

We take the fine-tuned Bob as our final model for evaluation. We also attempted an open-weight \texttt{Qwen3-4B-Thinking-2507} run with a LoRA adapter, which gave unstable training curves and was not pursued to evaluation (Appendix \ref{SEC:APPENDIX:QWEN3}).

\subsection{Intrinsic Evaluation}
\label{SEC:EXPERIMENTS:INTEVAL}

We first report how much Bob improves at the semantic inequivalence game itself after its single training round. Challenge instances are generated by the final Alice (round 7) from either MBPP-train source programs (as during training) or held-out MBPP-test programs, which neither agent has seen. With \texttt{gpt-4o-mini}, the percentage of instances solved by Bob rises from 75.99\% to \textbf{86.98\%} on MBPP-train sources and from 88.37\% to \textbf{91.67\%} on MBPP-test sources. The untrained Bob is already strong (we did not train Alice to convergence), yet a single Bob training round still substantially improves its ability to play the game, showing that our protocol effectively teaches LLMs to reason about the inputs on which program variants diverge.

\subsection{Extrinsic Evaluation}
\label{SEC:EXPERIMENTS:EXTEVAL}

Being proficient at playing the semantic inequivalence game may be directly useful in certain circumstances, such as determining whether a code refactoring has introduced subtle bugs.  
However, ultimately, we aim for this game to teach LLMs skills that generalize to a variety of tasks.  
Therefore, we evaluate our approach across a range of code-related tasks using standard benchmarks.

While we include code generation tasks, our primary focus is on code understanding.  
Therefore, we use the trained evaluator Bobs, denoted as \textbf{\tt sinq-gpt-4o-mini} (based on \texttt{gpt-4o-mini}), and \textbf{\tt sinq-gpt-4.1-nano} (based on \texttt{gpt-4.1-nano}) as our main checkpoints. Each trained model is primarily compared to its own base model.
We do not evaluate the models based on \texttt{Qwen3-4B-Thinking-2507}, as that set of experiments yielded inconsistent training curves and was aborted.

\subsubsection{Python Builtin Identifier Swap}
\label{SEC:EXPERIMENTS:EXTEVAL:IDSWAP}

The \textbf{Python builtin identifier swap} \citep{miceli-barone-etal-2023-larger} is a very challenging code-understanding benchmark. In its classification version, the model must decide which of two variants of a Python snippet is more likely correct, where each snippet is prepended with a statement reassigning two builtin functions, e.g.\ \texttt{print, len = len, print}. One variant is a function from a GitHub repository; the other is the same function with the two builtin identifiers (here \texttt{len} and \texttt{print}) swapped throughout. The reassignment makes the modified snippet correct but highly out-of-distribution, and the original in-distribution but incorrect. \citet{miceli-barone-etal-2023-larger} found this confused all state-of-the-art LLMs of the time, which even did worse as they grew larger, a case of \textbf{inverse scaling} \citep{mckenzie2023inverse}.

We evaluate \texttt{gpt-4o-mini} and \texttt{gpt-4.1-nano} (both released after the original study) and our trained Bobs \texttt{sinq-gpt-4o-mini} and \texttt{sinq-gpt-4.1-nano}, using either the original prompt or a chain-of-thought variant; results are in Table \ref{tab:experiments:exteval:idswap}.

\begin{table}[t]
    \centering
    \footnotesize
    \setlength{\tabcolsep}{6pt}
    \renewcommand{\arraystretch}{0.95}
    \begin{tabular}{lcccc}
        \toprule
        Base LLM & Baseline & \textbf{SInQ} & Baseline CoT & \textbf{SInQ CoT} \\
        \midrule
        gpt-4o-mini
        & 1.65 & \textbf{5.35} & 1.90 & \textbf{2.30} \\
        gpt-4.1-nano
        & \textbf{2.80} & 2.50 & \textbf{14.35} & 10.05 \\
        \midrule
        \multicolumn{5}{l}{\footnotesize\emph{Superseded: evaluator trained on instances validated by the flawed harness}} \\
        gpt-4.1-nano (old)
        & 2.80 & 7.40 & 14.35 & 15.30 \\
        \bottomrule
    \end{tabular}
    \caption{Accuracy on the Python builtin identifier swap benchmark for baseline (untrained Bob) and \textbf{SInQ} (trained Bob) models, with and without chain-of-thought prompting. The last row reports a superseded \texttt{gpt-4.1-nano} evaluator, trained before a bug in the verification harness was fixed; see the discussion below.}
    \label{tab:experiments:exteval:idswap}
\end{table}

Despite being released years after the benchmark, without chain-of-thought both base models still perform very poorly, worse even than GPT-3.5 (3.35\%)\footnote{Raw results for \citet{miceli-barone-etal-2023-larger} are available on the associated GitHub repository: \url{https://github.com/Avmb/inverse_scaling_prize_code_identifier_swap/blob/main/eval_chat_llms/eval_chat_llms_results.json}.}, confirming that it remains challenging. The two base models then behave differently: on \texttt{gpt-4o-mini} our approach improves accuracy without CoT (+3.7\%) and slightly with CoT (+0.4\%), whereas on \texttt{gpt-4.1-nano} it does not help, leaving accuracy essentially unchanged without CoT (-0.3\%) and lowering it with CoT (-4.3\%).

The last row of Table \ref{tab:experiments:exteval:idswap} reports a superseded \texttt{gpt-4.1-nano} evaluator that appears to improve substantially on this benchmark (+4.6\% without CoT, +0.95\% with CoT).\footnote{The version of this paper submitted for review reported this superseded evaluator in place of the corrected one, and consequently overstated the gains on this benchmark; the present version corrects it. The error is confined to the \texttt{gpt-4.1-nano} rows of this table: all other results in this paper, including every \texttt{gpt-4o-mini} result and all \texttt{gpt-4.1-nano} results on vulnerability detection (Section \ref{SEC:EXPERIMENTS:EXTEVAL:VULN}) and code generation (Section \ref{SEC:EXPERIMENTS:EXTEVAL:CODEGEN}), were obtained with the corrected evaluator and are unaffected.} That evaluator was trained on instances validated by an earlier version of our verification harness which contained a bug: it compared program outputs by Python object identity rather than by value, and failed to reject inputs invalid for the source program, so a fraction of its training instances were not in fact diverging. We report it for transparency rather than as a result: the same checkpoint performs \emph{worse} than the untrained model on both vulnerability-detection benchmarks (80.5\% versus 83.2\% on PySecDB under greedy decoding), so we do not read its apparent advantage here as evidence of generalizable skill.

This benchmark is quite different from our \textbf{semantic inequivalence game} training data, sharing only the need to reason about the semantics of unusual Python snippets. The \texttt{gpt-4o-mini} improvement therefore suggests that our approach can teach code-reasoning skills that transfer to quite distant tasks, but the \texttt{gpt-4.1-nano} results show that this transfer is not reliable across base models, and we do not claim it as a general property of the method.

We report additional results on this benchmark with state-of-the-art reasoning models in Appendix \ref{SEC:APPENDIX:ADD_IDSWAP}.

\subsubsection{Vulnerability Detection}
\label{SEC:EXPERIMENTS:EXTEVAL:VULN}

Security vulnerabilities often arise from counterintuitive behaviours, where a programmer's (human or LLM) understanding of the code's semantics differs from its actual behaviour on edge cases that evade pre-deployment testing. Our game incentivizes Alice to find exactly such edge cases and Bob to become robust to them: to win, Bob must not merely notice that a small edit \emph{exists} but construct an input on which it becomes observable, which is the reasoning vulnerability detection relies on, since a vulnerable program typically differs from its patched version only on a small set of edge-case inputs. A qualitative analysis of the edits Alice actually produces is given in Appendix \ref{SEC:APPENDIX:QUALEX}. We test this on two benchmarks.

\paragraph{PySecDB}
\citep{sun2023exploringsecuritycommitspython} is a dataset of Python commits, as diff patches, labelled by whether they contain a security fix. We ask the LLMs to classify each patch without the rest of the repository as context, making this challenging, and discard the few commits exceeding the 128,000-token context limit.

\begin{table}[t]
    \centering
    \footnotesize
    \setlength{\tabcolsep}{4pt}
    \renewcommand{\arraystretch}{0.95}
    \begin{tabular}{llcccccc}
        \toprule
        Benchmark & Base LLM 
        & Baseline & \textbf{SInQ} 
        & Baseline Maj & \textbf{SInQ Maj} 
        & Baseline CoT & \textbf{SInQ CoT} \\
        \midrule
        \multirow{2}{*}{PySecDB}
        & gpt-4o-mini  
        & 82.43 & 82.51 
        & 82.48 & \textbf{82.81} 
        & 73.55 & 73.00 \\
        & gpt-4.1-nano 
        & 83.20 & 83.33 
        & 83.03 & \textbf{83.35} 
        & 78.74 & 78.20 \\
        \midrule
        \multirow{2}{*}{CodeXGLUE}
        & gpt-4o-mini  
        & 55.23 & 55.60 
        & 55.12 & \textbf{56.04} 
        & 47.69 & 47.22 \\
        & gpt-4.1-nano 
        & 54.76 & \textbf{55.27} 
        & 54.83 & 55.23 
        & 52.20 & 49.85 \\
        \bottomrule
    \end{tabular}
    \caption{Vulnerability detection results on the PySecDB and CodeXGLUE Defect Detection benchmarks, with or without majority voting (out of 9 samples) or chain-of-thought prompting.}
    \label{tab:experiments:exteval:vuln:combined}
\end{table}


\paragraph{CodeXGLUE Defect Detection}
\citep{CodeXGlue2021} is a dataset of C/C++ snippets labelled by whether they contain known vulnerabilities, a particularly challenging target since our models were fine-tuned only on Python.


We use greedy classification (temperature 0.0), majority voting over 9 samples (temperature 1.0), and CoT (temperature 0.6, \(N = 1\)); results are in Table \ref{tab:experiments:exteval:vuln:combined}. Our approach yields small but consistent improvements across both datasets, tasks and languages, suggesting it confers additional vulnerability-reasoning ability despite no task-specific training.

We note that, for both the baseline and our models, chain-of-thought prompting \emph{lowers} accuracy on these benchmarks relative to direct classification, and \textbf{SInQ} degrades slightly under CoT on some model/dataset combinations. We attribute this to the nature of the task: these benchmarks require a holistic binary judgement over an entire diff or snippet rather than the construction of a single diverging input, and free-form reasoning appears to give the model more opportunities to argue itself out of a correct initial judgement. This mirrors the identifier-swap results (Section \ref{SEC:EXPERIMENTS:EXTEVAL:IDSWAP}), where what gains there are likewise occur without CoT, so the most consistent benefits of our method are obtained in the direct and majority-voting settings.

\subsubsection{Code Generation}
\label{SEC:EXPERIMENTS:EXTEVAL:CODEGEN}

We run a standard code generation experiment using the EvalPlus harness \citep{evalplus2023, evalperf2024}, which evaluates LLMs on the test portions of MBPP and HumanEval \citep{chen2021codex}, as well as on augmented versions of these datasets, MBPP+ and HumanEval+, which contain additional unit tests per instance; full Pass@1 rates are in Table \ref{tab:experiments:exteval:codegen} (Appendix \ref{SEC:APPENDIX:CODEGEN}). For \texttt{gpt-4o-mini} our approach improves Pass@1 on MBPP (+2.1\%) and MBPP+ (+0.8\%), matches it on HumanEval, and loses slightly on HumanEval+ (-0.6\%). For \texttt{gpt-4.1-nano} it gives a small drop on MBPP (-1.6\%) and MBPP+ (-0.3\%), is roughly unchanged on HumanEval (+0.6\%), and improves substantially on HumanEval+ (+3.1\%).

Although trained on MBPP-train \texttt{code}, our models were never trained on the MBPP task itself nor shown its natural-language instructions, and they are the evaluators (Bobs), not fine-tuned for code generation, yet they still improve or maintain generation performance. For generation-oriented tasks it may help to train a separate model combining the final Alice and Bob datasets.

\section{Discussion}
\label{SEC:DISCUSSION}

\paragraph{A neurosymbolic reading.}
Our gains are largest on code \emph{understanding} and neutral on generation, as expected since we train only the evaluator. The design choice behind them is that the learning signal is computed symbolically rather than learned: every round is adjudicated by executing both programs on a candidate input under the operational semantics of the language, so a diverging input is a \emph{certificate} of inequivalence, checkable in bounded time, with no reward model, LLM judge or human labels in the loop. SInQ therefore sits in the neurosymbolic tradition of supervising a neural learner with a symbolic verifier, and the asymmetry it exploits is itself formal: deciding full semantic equivalence is impossible in general by Rice's theorem (Appendix \ref{SEC:APPENDIX:NDSINEQ}), but inequivalence is semi-decidable and its witnesses are cheap to check, so we build the game on the side of the problem where symbolic verification is tractable and delegate to the neural component the search that verification cannot perform. This is also why the game has no theoretical performance cap: the verifier stays sound however strong the players become, whereas a learned reward can be gamed.

\paragraph{Relation to other self-play approaches.}
\citet{zhao2025absolutezeroreinforcedselfplay} also apply self-play to code with an executor for validation, but their proposer invents tasks from scratch, whereas our generator is \emph{grounded} in a corpus of real-world programs, which we regard as the main reason the skill transfers to real security commits and to a language unseen in training. \citet{dong2025stp} apply a conjecture-and-prove variant to formal theorem proving, where the verifier is a proof assistant and the statements are formal by construction; our contribution in that comparison is to show that an ordinary interpreter can serve as the verifier for informal, real-world code. Subsequent work by \citet{nok-barone-2026-improving} instead strengthens the verifier, using Liquid Haskell proofs as certificates of \emph{equivalence} while keeping execution-based counterexamples for inequivalence; their ablations indicate that inequivalence supervision of the kind we use here mainly contributes data volume, while the equivalence proofs account for most of the gain in reasoning ability. The strength of the symbolic verifier, rather than the self-play loop by itself, therefore looks like the most promising axis for improvement, at the price of restricting the method to languages with mature verification tooling.

\section{Conclusions}
\label{SEC:CONCLUSIONS}

We presented a self-play method, the \textbf{semantic inequivalence game}, for improving the code-understanding capabilities of LLMs, which is both \textbf{grounded} in a dataset of real-world programs and free of any \textbf{theoretical performance cap}. Evaluated on two vulnerability-detection benchmarks, it learns skills that generalize across tasks and programming languages, and on the challenging \textbf{Python builtin identifier swap} benchmark it substantially improves one of the two base models we fine-tune. We release all code and the generated synthetic data; exact replication, up to sampling randomness, should be possible with a modest budget (approximately \$600) as long as \texttt{gpt-4o-mini-2024-07-18} and \texttt{gpt-4.1-nano-2025-04-14} remain available on the OpenAI platform.

\section*{Limitations}
\label{SEC:LIMITATIONS}

Our method has the following limitations, primarily due to our limited budget.
\textbf{Model scale:} we fine-tuned only \texttt{gpt-4o-mini} and \texttt{gpt-4.1-nano}, plus an unstable open-weight \texttt{Qwen3-4B-Thinking-2507} run (Appendix \ref{SEC:APPENDIX:QWEN3}); larger and state-of-the-art reasoning models would strengthen the empirical claims.
\textbf{Training regime:} we used supervised fine-tuning on the OpenAI platform, likely with LoRA-style adapters; reinforcement learning and full-parameter tuning remain to be explored.
\textbf{Self-play depth:} due to budget, Bob was trained for a single round; multiple rounds with Alice trained to convergence between them, as in AlphaZero-style self-play, are left to future work.
\textbf{Baselines:} we do not compare against non-adversarial synthetic data (e.g.\ rule-based perturbations or hardness-filtered examples), which would isolate the benefit of self-play. We expect a fixed perturbation scheme to plateau once its patterns are learned, whereas our game has no intrinsic performance cap (Appendix \ref{SEC:APPENDIX:NDSINEQ}); confirming this remains future work.

\acks{\textbf{Antonio Valerio Miceli Barone} has received funding from a collaboration between the University of Edinburgh and Cisco Systems led by \textbf{Vaishak Belle} (University of Edinburgh) and \textbf{Ali Payani} (Cisco Systems).

Additional funding was provided by the \textbf{UKRI Engineering and Physical Sciences Research Council} and the \textbf{Artificial Intelligence Security Institute} under the grant ``\textbf{Understanding and Improving the Behaviour of AI Agents in Competitive and Cooperative Games}'', and by Amazon through the \textbf{AWS Agentic Amazon Research Award} titled ``\textbf{Diffusion-inspired chain-of-thought self-revision}''.}

%


\bibliography{custom}

\appendix

\section{SInQ training algorithm}
\label{SEC:APPENDIX:ALGORITHM}

\begin{algorithm2e}[h]
\caption{SInQ training with rejection-sampling SFT and difficulty targeting.}
\label{ALG:SINQ}
\DontPrintSemicolon
\KwIn{source programs $\mathcal{S}$; base LLM $M_0$; samples per query $N$; hard threshold $\tau = 5$}
\KwOut{trained evaluator Bob}
Alice $\gets M_0$;\quad Bob $\gets M_0$;\quad $\mathcal{D}_A \gets \emptyset$\;
\Repeat{Bob stops improving}{
  \Repeat{mean difficulty converges}{
    \ForEach{$P \in \mathcal{S}$}{
      sample $O \sim$ Alice$(\mathit{Template}_A(P, d_t{=}10))$\;
      extract $(\mathit{CoT}, Q, x)$ from $O$\;
      \lIf{$O$ invalid \textbf{or} $P(x) = Q(x)$}{discard, \textbf{continue}}
      $d \gets 10\,(1 - N_\mathrm{correct}/N)$ by sampling Bob $N$ times on $(P,Q)$\;
      add $(\mathit{Template}_A(P, d),\, O)$ to $\mathcal{D}_A$\;
    }
    bias $\mathcal{D}_A$ towards hard ($d \ge \tau$) instances and add difficulty-prediction examples\;
    Alice $\gets$ SFT$(M_0, \mathcal{D}_A)$ \tcp*{retrain from base}
  }
  collect Bob's correct plays $(P, Q, \hat{x})$ into $\mathcal{D}_B$\;
  Bob $\gets$ SFT$(M_0, \mathcal{D}_B)$\;
  recompute the difficulty of every instance in $\mathcal{D}_A$ against the new Bob\;
}
\Return Bob\;
\end{algorithm2e}

\section{Ethical Considerations}
\label{SEC:ETHICS}

Our proposed method involves training LLMs on synthetically generated data based on existing open-source programming code datasets.
We also evaluate our method on open-source datasets.
No human experiments were conducted, and therefore, the risk that our experiments have caused harm to individuals or infringed upon anyone's intellectual property rights is negligible.

Our work aims to enhance LLMs' ability to reason about programming code.
There is a potential risk that such capabilities could be used for unethical activities, such as hacking computer systems. However, these capabilities can also be used to strengthen computing systems by detecting security vulnerabilities in codebases.
We believe that the net societal impact of our research will be positive.

\section{Open-weight \texttt{Qwen3-4B-Thinking} experiments}
\label{SEC:APPENDIX:QWEN3}

\begin{figure}[ht]
\begin{center}
\includegraphics[width=0.48\linewidth]{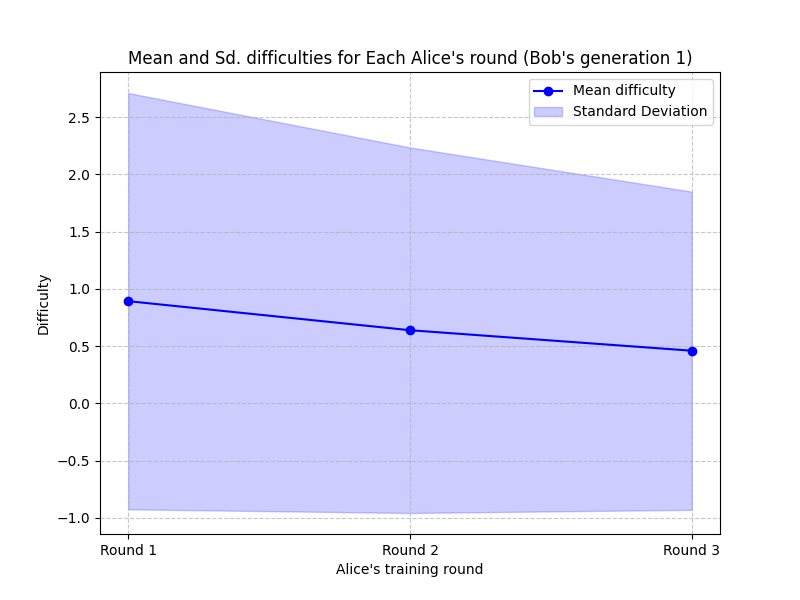}
\includegraphics[width=0.48\linewidth]{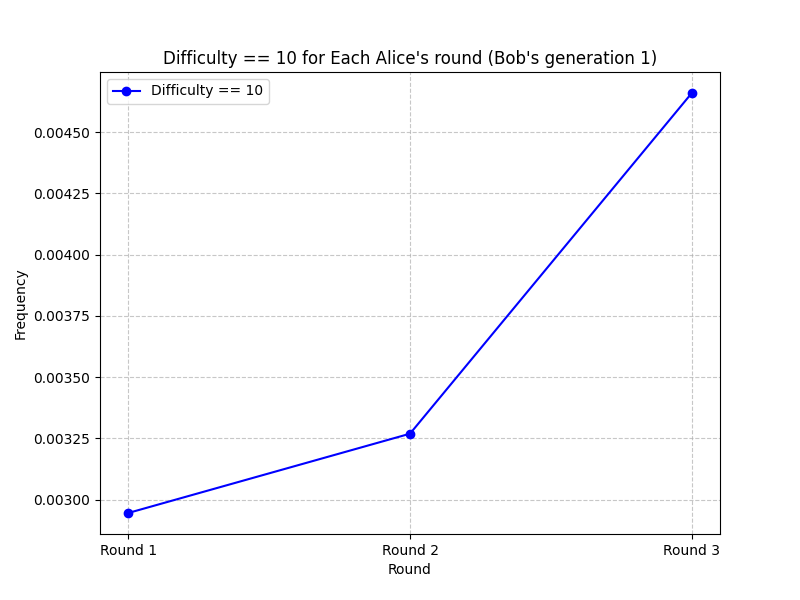}
\end{center}
\caption{Instance difficulty with respect to an untrained Bob, plotted against the number of Alice's training rounds, \texttt{Qwen3-4B-Thinking-2507} models. Left: mean and standard deviation. Right: fraction of instances with maximum difficulty.}
\label{FIG:EXPERIMENTS:TRAIN:QWEN3_4B_THINKING_DIFF_CURVES}
\end{figure}

We also attempted a set of experiments based on the Alibaba Qwen3 Thinking LLM, specifically the smallest non-quantized version available on Hugging Face \texttt{Qwen3-4B-Thinking-2507}.
We performed generation using the recommended hyper-parameters, with full context length ($32768$ tokens).
We used a rank-stabilized LoRA adapter with rank = $128$, alpha = $16$, dropout probability = $0.1$, epochs = $5$, maximum learning rate = $2e-4$.
We did not use difficulty prediction examples, since according to the Qwen3 fine-tuning guidelines, CoT traces should not be included in assistant messages in multi-turn conversations except on the last one.
This set of experiments led to unstable results, with the proportion of examples with maximum difficulty increasing with each of Alice's rounds, as expected, but the mean difficulty decreasing (training curves are reported in Figure \ref{FIG:EXPERIMENTS:TRAIN:QWEN3_4B_THINKING_DIFF_CURVES}).
The two metrics moving in opposite directions indicates that Alice learns to produce a growing fraction of maximally hard instances while the bulk of its generations become \emph{easier}, rather than the whole difficulty distribution shifting upwards as it does for the OpenAI models.
We interpret this as a plateau or overfitting effect of the small base model and low-rank adapter, which can latch onto a few hard patterns without improving its generation distribution overall, rather than evidence of a fundamental difference between the OpenAI and Qwen3 model families.
Because most of our budget consisted of OpenAI API credits, we could only run this open-model experiment on the smallest Qwen3 reasoning model and explore a small set of fine-tuning hyperparameters; we leave a more thorough open-model study, ideally with larger models and a wider hyperparameter search, to future work.
We did not evaluate these models on the downstream benchmarks, as the training curves were inconsistent.

\section{Difficulty-prediction training examples}
\label{SEC:APPENDIX:DIFFPRED}

In addition to the main generation examples, we explicitly train Alice to predict instance difficulty by including training examples in the following format:
\begin{equation*}
\begin{split}
    TE_{diff} := (& Template_{Alice}(P, \text{"Any"}), \\
                  & O, \\
                  & Template_{diff_{in}}, \\
                  & Template_{diff_{out}}(d(P, Q, Bob)))
\end{split}
\end{equation*}
where the target difficulty in the first "user" message is replaced by the string "Any", and the first "assistant" message contains Alice's self-generated output instance. The second "user" message provides an instruction to predict the difficulty of the instance, and the second "assistant" message contains the actual difficulty. We minimize the loss only on the second "assistant" message.
This part of the dataset is also biased towards hard examples, but we set the number of easy examples at 50\%, as we are not minimizing the loss on the tokens of easy instances but only on their difficulty prediction. Therefore, including these examples is unlikely to be detrimental.

\section{A worked example of the game}
\label{SEC:APPENDIX:EXAMPLE}

The following is an artificial example of a single round of the semantic inequivalence game (see Appendix \ref{SEC:APPENDIX:QUALEX} for a real example from the MBPP dataset).

\noindent 1. Alice receives program $P$:
\begin{lstlisting}[language=Python, basicstyle=\footnotesize]
def fib(n):
    if n <= 0:
        return 0
    elif n == 1:
        return 1
    return fib(n - 1) + fib(n - 2)
\end{lstlisting}

and returns program $Q$:
\begin{lstlisting}[language=Python, basicstyle=\footnotesize]
def fib(n):
    if n == 0:
        return 0
    elif n == 1:
        return 1
    return fib(n - 1) + fib(n - 2)
\end{lstlisting}
together with diverging input $x$:
\begin{lstlisting}[language=Python, basicstyle=\footnotesize]
{"n": -1}
\end{lstlisting}

\noindent 2. Both programs are run in a sandbox with input $x$, which results in $P$ returning $0$ while $Q$ recurs until it triggers either the recursion limit of the Python interpreter or the randomized time limit, proving that $x$ is indeed a diverging input.

\noindent 3. Bob receives both $P$ and $Q$ and generates a possibly different diverging input \(\hat{x}\), e.g.:
\begin{lstlisting}[language=Python, basicstyle=\footnotesize]
{"n": -2}
\end{lstlisting}
$P$ and $Q$ are evaluated on input \(\hat{x}\), proving that this is also a diverging input; therefore, Bob wins this round.

\newpage

\section{Qualitative analysis}
\label{SEC:APPENDIX:QUALEX}

We manually analysed some of Alice's outputs in order to understand what kind of manipulation it introduces. A typical pattern is the introduction of subtle bugs in conditional branches that deal with edge cases, of the sort that often occur in programs with security vulnerabilities. Winning the game requires Bob to reason about which edge-case inputs propagate such a change into an observable difference in behaviour, and identifying exactly those inputs is what distinguishes a real vulnerability from a benign refactoring; this could explain the improvements we observe on the vulnerability datasets (Section \ref{SEC:EXPERIMENTS:EXTEVAL:VULN}), despite our models never being trained on vulnerability data.

The listing below is an MBPP implementation of maximum-subarray-sum used as a source program. Alice's variant is identical except that the highlighted \verb|max_ending_here = 0| statement is removed (and the entry-point function renamed), an easy-to-miss change that causes it to mishandle negative array values.
\begin{lstlisting}[language=Python, escapechar=!, basicstyle=\footnotesize]
from sys import maxsize

def max_sub_array_sum(a,size):
    max_so_far = -maxsize - 1
    max_ending_here = 0
    start = 0
    end = 0
    s = 0
    for i in range(0,size):
        max_ending_here += a[i]
        if max_so_far < max_ending_here:
            max_so_far = max_ending_here
            start = s
            end = i
        if max_ending_here < 0:
            !\colorbox{green}{\texttt{max\_ending\_here = 0}}!
            s = i+1
    return (end - start + 1)
\end{lstlisting}

\section{\texttt{gpt-4.1-nano} training curves}
\label{SEC:APPENDIX:NANOCURVES}

\begin{figure}[ht]
\begin{center}
\includegraphics[width=0.48\linewidth]{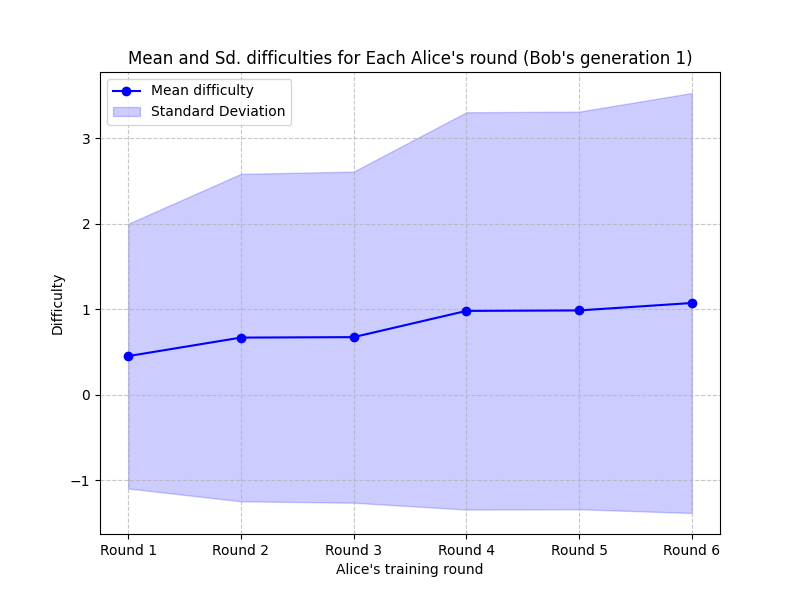}
\includegraphics[width=0.48\linewidth]{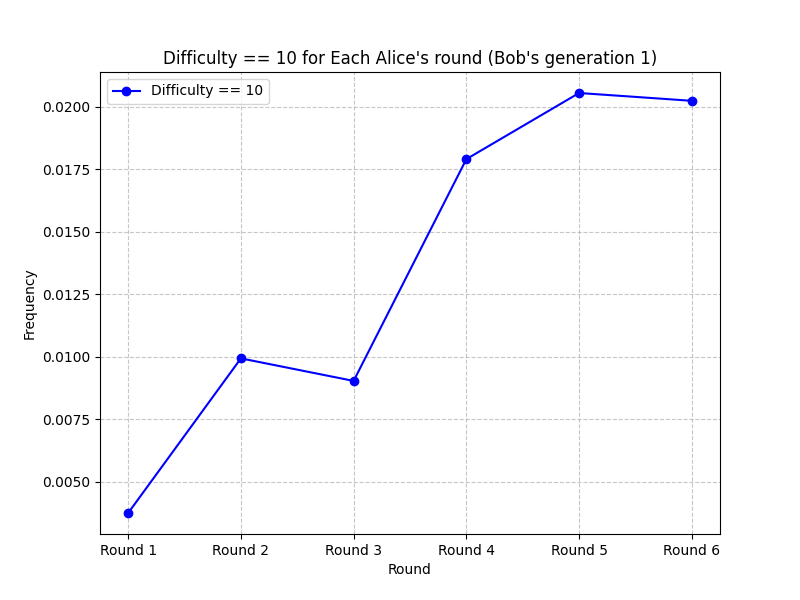}
\end{center}
\caption{Instance difficulty with respect to an untrained Bob, plotted against the number of Alice's training rounds, \texttt{gpt-4.1-nano} models. Left: mean and standard deviation. Right: fraction of instances with maximum difficulty.}
\label{FIG:EXPERIMENTS:TRAIN:GPT_4_1_NANO_DIFF_CURVES}
\end{figure}

\section{Generation and validation details}
\label{SEC:APPENDIX:EXPDETAILS}

Both agents output markdown, using sections to separate their CoT traces from the final outputs embedded in Python code blocks. Outputs are extracted, syntactically validated and normalized with the Mistune markdown parser\footnote{\url{https://mistune.lepture.com/en/latest/}} and the Python \texttt{ast} parser/unparser (parsing and unparsing also strips comments and unusual indentation, blocking trivial non-semantic attacks). We then check each diverging input by executing both programs in a test harness under a randomized time limit, uniformly sampled in \([2.5, 5.5]\)\,s to discourage instances relying on race conditions.

\section{Non-decidability of semantic inequivalence}
\label{SEC:APPENDIX:NDSINEQ}
In a semantic equivalence game where the task of the evaluator (``Bob'') is to determine whether two programs $P$ and $Q$ are equivalent, there is clear undecidability due to a trivial consequence of Rice's theorem \citep{Rice:53}.
Since a perfect Bob cannot exist, this results in a ``full-employment theorem'' \citep{AppelModernCompilerImpl1998} for Alice: in principle, it can always find new ways to fool Bob.
This iterative process leads to increasingly stronger Bobs, who in turn train progressively stronger Alices.  

However, in the semantic inequivalence game (section \ref{SEC:METHOD:INEQ}), the programs $P$ and $Q$ given to Bob are guaranteed not to be equivalent, and Bob's task is to find a diverging input $\hat{x}$ such that $P(\hat{x}) \neq Q(\hat{x})$, which is guaranteed to exist.
It may be asked whether this constraint on the programs makes the problem any logically easier.
We show here that this is not the case.

\subsection*{Definitions}
\label{SEC:APPENDIX:NDSINEQ:DEFS}
Given an arbitrary, but fixed, admissible numbering (programming language) of partial computable functions, we define a program $P$ as an index in such numbering.\\
With a slight overload of notation, we denote $P(x)$ as the result of evaluating on input $x$ the partial computable function defined by program $P$.
Without loss of generality, we consider the inputs of our programs to be the natural numbers and the outputs to be natural numbers plus the special value $\bot$ that denotes non-termination.\\

The mapping between programs and functions is surjective but not injective: each function can be defined by infinitely many programs.
We define two programs $P$ and $Q$ \textbf{equivalent} if they define the same function, conversely we define them \textbf{inequivalent} if they define different functions, that is, if there exists at least one \textbf{diverging input} $\hat{x}$ such that $P(\hat{x}) \neq Q(\hat{x})$.

Given a program $A$ and a natural number $n$, the \textbf{halting problem}, denoted by the predicate $Halt(A, n)$, consists of determining whether $A(n) \neq \bot$, which is notoriously undecidable in the general case.

\begin{theorem}
\label{SEC:APPENDIX:NDSINEQ:THEOREM_1}
There is no perfect evaluator program $\overset{*}{Bob}$ such that, for any inequivalent programs $P$ and $Q$ it computes a diverging input for $P$ and $Q$.
\end{theorem}
\begin{proof}
If programs $P$ and $Q$ have a diverging input for which they both halt producing distinct output values: $P(\hat{x}) = y_p \in \mathbb{N}$,  $Q(\hat{x}) = y_q \in \mathbb{N}$ and $y_p \neq y_q$, then $\overset{*}{Bob}$ can compute $\hat{x}$ by dovetailing.
The interesting case is when for each diverging input only one between $P$ and $Q$ halts.
We show that such diverging inputs cannot be computed in the general case by a reduction to the halting problem.

Given a program $A$ and a natural number $n$, it is possible to algorithmically construct two programs $P_{A,n}^*$ and $Q_{A,n}^*$ defined as follows:
\begin{lstlisting}[language=Python, basicstyle=\footnotesize, caption=Definition of $P_{A,n}^*$ and $Q_{A,n}^*$]
def P_A_n_star(x):
    if x == 0:
        A(n)
        return 1
    else:
        while True:
            pass

def Q_A_n_star(x):
    if x == 0:
        return 1
    else:
        P_A_n_star(x-1)
\end{lstlisting}

By construction, if $A$ halts on input $n$, then $P_{A,n}^*$ and $Q_{A,n}^*$ diverge only on input $1$:\\ 
$A(n) \neq \bot \iff P_{A,n}^*(1)=\bot \wedge Q_{A,n}^*(1)=1$,\\ 
otherwise if $A$ does not halt on $n$, then $P_{A,n}^*$ and $Q_{A,n}^*$ diverge only on input $0$:\\ 
$A(n) = \bot \iff P_{A,n}^*(0)=\bot \wedge Q_{A,n}^*(0)=1$.\\
They are always equivalent on any other input.
Therefore:
\begin{lstlisting}[language=Python, basicstyle=\footnotesize, caption=Halting decider]
def Halt(A, n):
    def P_A_n_star(x): ... 
    # defined as in Listing 1
    
    def Q_A_n_star(x): ... 
    # defined as in Listing 1
    
    return Bob_star(P_A_n_star, 
                    Q_A_n_star) == 1
\end{lstlisting}
Since a general program that decides the halting problem cannot exist, then a perfect evaluator for the semantic inequivalence problem $\overset{*}{Bob}$ cannot exist.
\end{proof}

It can be noted that the proof of Theorem \ref{SEC:APPENDIX:NDSINEQ:THEOREM_1} applies in the general case but deviates from the constraints of the semantic inequivalence game in two important aspects: 
\begin{enumerate}
    \item The proof involves distinguishing between the halting behaviour of programs under arbitrary runtime, while in the game programs are checked against the diverging inputs produced by Alice and Bob under a time limit, after which they are assumed to return a special "TIMEOUT" value.
    \item In the proof we allow both Bob's input programs $P$ and $Q$ to take a special form that depends on the program $A$ whose halting behaviour is under consideration, while in the game the program $P$ is sampled from a dataset and Alice only controls program $Q$.
\end{enumerate}
It may be asked whether these constraints make Bob's task substantially easier, allowing for a perfect $\overset{*}{Bob}$ to exist, which would imply a performance cap.
We show that this is not the case.

In order to address the first point, we note that while the halting problem under a time limit is decidable if we only require the halting detector program to eventually halt, it is still undecidable if the halting detector program has to halt itself within the same time limit of the program it checks\footnote{This is provable with an argument about program length.}.
Therefore, by constraining Bob's resource usage, allowing Alice to always have more resources than Bob, and gradually increasing the time limit of the programs, it is possible for Alice to always generate harder and harder instances.
Once Bob stops learning, the resource limits can be increased, enabling further learning, in principle forever.
In practice, Alice and Bob are implemented as agents based on LLMs operating in chain-of-thought mode, thus resource limits can be enforced by controlling the number of reasoning tokens, or in the long term by controlling the parameter count, layer count, or expert count of the base LLMs\footnote{Assuming that LLMs always become better at learning when increasing their resource limits.}.

As for the second point, we show that for any \textbf{non-trivial program} $P$, Alice can generate a program $\overline{Q_{P,A,n}}$ which checks whether program $A$ halts on input $n$, where by ``non-trivial'' we mean that there exist at least two distinct inputs $x_0$ and $x_1$ such that $P$ halts on both, returning two distinct values, respectively $y_0$ and $y_1$:

\begin{lstlisting}[language=Python, basicstyle=\footnotesize, caption=Definition of $\overline{Q_{P,A,n}}$]
def Q_P_A_n_bar(x):
    if (x == x_0) or (x == x_1):
        if Halt(A, n):  # defined as in Listing 2
            return y_0
        else:
            return y_1
    else:
        return P(x)
\end{lstlisting}
This is a self-referential construction, where Bob is tasked to analyse a program that invokes Bob itself, thus Bob has to analyse its own behaviour.
If Bob was indeed the perfect $\overset{*}{Bob}$, then $P$ and $\overline{Q_{P,A,n}}$ would return different values only on input $x_0$ if $A$ halts on $n$, or only on $x_1$ if it does not, thus solving the halting problem.
Note that this construction is still a valid output for Alice even when Bob is not perfect, since $\overline{Q_{P,A,n}}$ will still differ from $P$ on $x_0$ or $x_1$ (possibly on both if the inner call to Bob does not halt), which means that in principle Alice can generate hard examples for Bob from arbitrary source programs, as long as they meet minimal ``non-triviality'' conditions.
In practice, we want the generated programs to run quickly on the CPU without invoking LLMs, so this self-referential construction is unwieldy, but it serves as a proof of concept which shows that arbitrarily complex logic can be added by Alice in the programs it generates, even starting from minimally complex source programs.

\section{Setting a target difficulty}
\label{SEC:APPENDIX:TARGETDIFFICULTY}

In the implementation of the semantic inequivalence game which we use in our experiment, we instruct the generator ``Alice'' to create challenge instances for the evaluator ``Bob'' with a specific target difficulty, defined as $10$ times the probability that Bob fails to solve the instance when invoked in sampling mode.
Setting the target difficulty always at the maximum value of $10$ makes the game equivalent to its original formulation in section \ref{SEC:METHOD:INEQ}, which, if Alice never produces invalid instances, is a \textbf{zero-sum game}.

It may be asked whether this maximally adversarial setting is always ideal.
Consider the following Python program that Alice may potentially generate:

\begin{lstlisting}[language=Python, basicstyle=\footnotesize, caption=Cryptographically hard $Q$ generated for a given $P$]
import hashlib

def Q(x):
  try:
    e = str(x).encode("utf-8")
    h = hashlib.sha3_256(e).hexdigest()
    t = ("af9ac3dac56b02f1ea017e7657a9bb7e" 
         "1778274e31509f134f023e41a5953866")
    if h == t:
      return "Bananas"
  except:
    pass
  return P(x)
\end{lstlisting}

For inputs $x$ that have the specific SHA-3-256 value defined in the code, $Q$ returns the string \texttt{Bananas}, otherwise it behaves as $P$; therefore, as long as $P$ does not happen to also return \texttt{Bananas} for all these specific inputs, they are diverging inputs.

Alice can easily generate this instance by first choosing a diverging input $x$ (in this example, the string ``correct horse battery staple''), then hashing it and hardcoding its hash value into $Q$, but, in order to solve this instance Bob has to successfully execute a \textbf{preimage attack} on SHA-3-256, which is considered a strong cryptographic function \citep{sha3}.
While this attack is theoretically possible by brute-force search, in practice it would require a runtime longer than the age of the universe, unless perhaps Bob is a cryptanalysis genius and manages to find a serious flaw in SHA-3-256, and even in this case, if the \textbf{one-way function conjecture} happens to be true then it is possible to construct asymptotically strong cryptographic hash functions \citep{levin2003taleonewayfunctions}, making Bob's task effectively hopeless.

The construction used in our specific example would require Alice to run code in order to compute the hash of its chosen diverging input, which current LLMs are typically not allowed to do in their default configuration and not in our experiments (although some common ``LLM agent'' setups do allow it), but Alice could still manage to create cryptographic puzzles which are too hard for any practical Bob to solve.

If Alice is instructed to always generate maximally difficult instances, it has an incentive to generate cryptographic puzzles, but since Bob only learns from the instances it can actually solve, this would effectively cause the learning process to stall.
In Appendix \ref{SEC:APPENDIX:NDSINEQ} we have proven that learning can continue forever in the limit of infinite computing resources, but in reality computing resources are finite, and cryptographic puzzles could stop the learning process as soon as Alice discovers the trick.
Even if it never resorts to cryptographic puzzles, Alice could just learn faster than Bob, eventually overwhelming Bob with instances that it cannot solve and thus stopping the learning process.

Fortunately, we can avoid this problem completely by setting the target difficulty to a value lower than the maximum, e.g. $7$, corresponding to the current Bob solving the instances with $30\%$ probability.
This changes the nature of the game from \textbf{zero-sum} to \textbf{positive-sum}, where Alice acts as a teacher that challenges Bob with instances which are hard, but not too hard for its current level.
As Bob improves, the difficulty of a given distribution of instances decreases, which in turn causes Alice to learn to recalibrate its difficulty estimation and gradually generate more challenging instances, enabling the training process to continue learning interesting coding logic for as long as the capacity of the underlying LLMs is not exceeded.

In our experiments, due to our limited resources, we could not train Alice to the point that it could seriously challenge Bob, thus we always set the target difficulty to $10$, but as a training recipe, we do recommend reducing the target difficulty if at some point Bob starts to fall behind.

\section{Prompt templates}
\label{SEC:APPENDIX:TEMPLATES}

\paragraph{System prompt for Alice}
{\footnotesize
\begin{spverbatim}
You are an expert computer scientist. Your task is to take a Python 3.10 program and write a similar program which is not semantically equivalent, which means that there must exist at least a diverging input example such that the original program and your program either produce different outputs or exceptions, or one halts and the other one does not halt. In addition to a program, you need to produce a diverging input example. Start by carefully analyzing the original program and think of how an example would propagate through it from the input to the return value, considering how to modify the program in order to elicit a different behavior. Make sure that the return values or exceptions raised by your program are picklable.
The original program and your program will be used in a test to evaluate the skill of an expert computer scientist who will have to produce a diverging example (not necessarily the same as yours), so make sure that the difference you introduce are not very easy to understand. You will be given a difficulty level from 0 (easiest) to 10 (hardest) to target. E.g. difficulty level 0 means that an expert computer scientist in the bottom decile or above should be able to find a diverging example, difficulty level 9 means that only an expert computer scientist in the top decile should be able to find a diverging example, and difficulty level 10 means that only the top 1\% or less of expert computer scientists should be able to find a diverging example.
Think step by step before writing your program. Use the following Markdown format, making sure that the following sections are delimited by level 1 headings, since they will have to be automatically parsed:
# Analysis
step by step analysis. This section can include sub-headings and code blocks
# Generated program
your program inside a Python code block. Do not change the name or signature of the entry point function
# Diverging input example
your diverging input example as a Python dictionary inside a Python code block
For instance, if the entry point function takes two parameters a and b and your diverging example is a="foo" and b=42, write:
```python
{
  "a": "foo",
  "b": 42
}
```
do not write the expected outputs
\end{spverbatim}
}

\newpage
\paragraph{User message for Alice}
As a Python f-string:

{\footnotesize
\begin{verbatim}
f"""Difficulty level: {difficulty_level}
Entry point function: {function_name}

```python
{code}
```"""
\end{verbatim}
}

During inference {\tt difficulty\_level} is the target difficulty (always $10$), during SFT training, for Alice's main examples it is the measured difficulty approximated to the nearest integer, for Alice's difficulty prediction examples it is the string {\tt "Any"}.

\paragraph{Second user message for Alice}
Used only for the difficulty prediction training examples.

{\footnotesize
\begin{spverbatim}
Predict the difficulty level of the instance. Just write "Difficulty level: D" where D is your prediction, do not write anything else.
\end{spverbatim}
}

\paragraph{Second assistant message for Alice}
Used only for the difficulty prediction training examples.
As a Python f-string:

{\footnotesize
\begin{verbatim}
f"""Difficulty level: {difficulty_level}"""
\end{verbatim}
}

where {\tt difficulty\_level} is the measured difficulty.

\paragraph{System prompt for Bob}

{\footnotesize
\begin{spverbatim}
You are an expert computer scientist. Your task is to take two Python 3.10 programs and determine whether or not they are semantically equivalent. Two programs are semantically equivalent if there exists no diverging input examples such that the original program and your program either produce different outputs or exceptions, or one halts and the other one does not halt. If you determine that the two programs are not semantically equivalent, you also need to produce a diverging input example. Start by carefully analyzing the two programs and think of how an example would propagate through them from the input to the return value, considering whether it could elicit a different behaviors. 
Think step by step before writing your program. Use the following Markdown format, making sure that the following sections are delimited by level 1 headings, since they will have to be automatically parsed:
# Analysis
step by step analysis. This section can include sub-headings and code blocks
# Equivalent?
Yes or No
# Diverging input example
your diverging input example as a Python dictionary inside a Python code block, or nothing if the two programs are equivalent.
For instance, if the entry point function takes two parameters a and b and your diverging example is a="foo" and b=42, write:
```python
{
  "a": "foo",
  "b": 42
}
```
do not write the expected outputs
\end{spverbatim}
}

Note that we ask Bob to determine whether the two programs are equivalent, even though they never are.
This is not strictly necessary, but it potentially makes the task slightly more difficult for Bob, which is beneficial since Bob tends to be much stronger than Alice.

\paragraph{User message for Bob}
As a Python f-string:

{\footnotesize
\begin{verbatim}
f"""Entry point function: {function_name}

Program 1:
```python
{code_P}
```

Program 2:
```python
{code_Q}
```"""
\end{verbatim}
}

The evaluation prompts will be included in the code released upon publication.

\section{Code generation results}
\label{SEC:APPENDIX:CODEGEN}

Full Pass@1 rates for the EvalPlus code generation experiment of Section \ref{SEC:EXPERIMENTS:EXTEVAL:CODEGEN}.

\begin{table}[ht]
    \centering
    \footnotesize
    \setlength{\tabcolsep}{4pt}
    \renewcommand{\arraystretch}{0.95}
    \begin{tabular}{lcccc}
        \toprule
        \multirow{2}{*}{Test set}
        & \multicolumn{4}{c}{Base LLM} \\
        \cmidrule(lr){2-5}
        & \multicolumn{2}{c}{gpt-4o-mini}
        & \multicolumn{2}{c}{gpt-4.1-nano} \\
        \cmidrule(lr){2-3} \cmidrule(lr){4-5}
        & Baseline & \textbf{SInQ}
        & Baseline & \textbf{SInQ} \\
        \midrule
        MBPP        & 82.8 & \textbf{84.9} & \textbf{87.6} & 86.0 \\
        MBPP+       & 69.6 & \textbf{70.4} & \textbf{72.5} & 72.2 \\
        HumanEval   & \textbf{87.2} & \textbf{87.2} & 89.6 & \textbf{90.2} \\
        HumanEval+  & \textbf{82.9} & 82.3 & 84.1 & \textbf{87.2} \\
        \bottomrule
    \end{tabular}
    \caption{Pass@1 rates (in \%) on the EvalPlus suite for baseline (untrained Bob) and \textbf{SInQ} (trained Bob) models.}
    \label{tab:experiments:exteval:codegen}
\end{table}

\section{Additional Python builtin identifier swap results}
\label{SEC:APPENDIX:ADD_IDSWAP}

\begin{figure}[ht]
    \centering
    \includegraphics[width=0.95\linewidth]{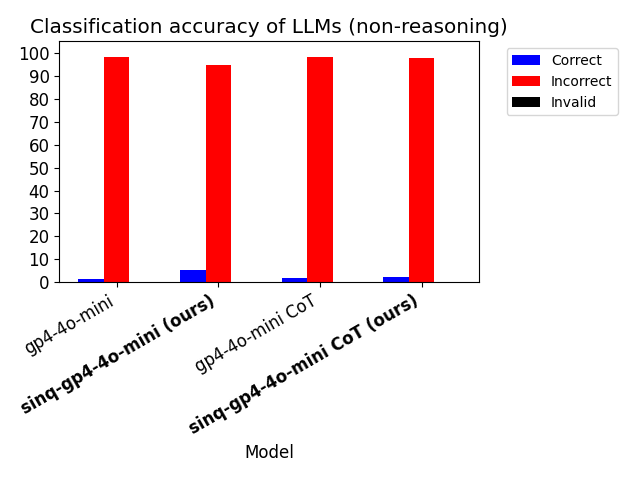}
    \caption{Python builtin identifier swap results for the baseline {\tt gpt-4o-mini} (untrained Bob) and our model {\tt sinq-gpt-4o-mini} (trained Bob), with or without chain-of-thought.}
    \label{FIG:APPENDIX:ADD_IDSWAP_NON_REASONING}
\end{figure}

\paragraph{Main results}
The \texttt{gpt-4o-mini} rows of Table \ref{tab:experiments:exteval:idswap}, presented as a bar chart in Figure \ref{FIG:APPENDIX:ADD_IDSWAP_NON_REASONING}.

\paragraph{Results on Reasoning Models}
Large Reasoning Models (LRMs) are LLMs which have been specifically trained to solve reasoning tasks, primarily in the domains of math and coding, using Chain-of-Thought reasoning.
These models, such as OpenAI {\tt o1} and {\tt o3} and DeepSeek-r1 \citep{deepseekai2025deepseekr1incentivizingreasoningcapability} typically generate a large amount of reasoning tokens during inference, hence they are said to perform \textbf{inference-time scaling} by trading off speed and cost for quality.
In practice, they are very strong but also very expensive.

Our approach could be broadly considered a type of LRM, since it is trained to solve reasoning problems using CoT, though in practice we use a much smaller base LLM and invest far fewer resources, both at training and at inference time.

\begin{figure}[ht]
    \centering
    \includegraphics[width=0.95\linewidth]{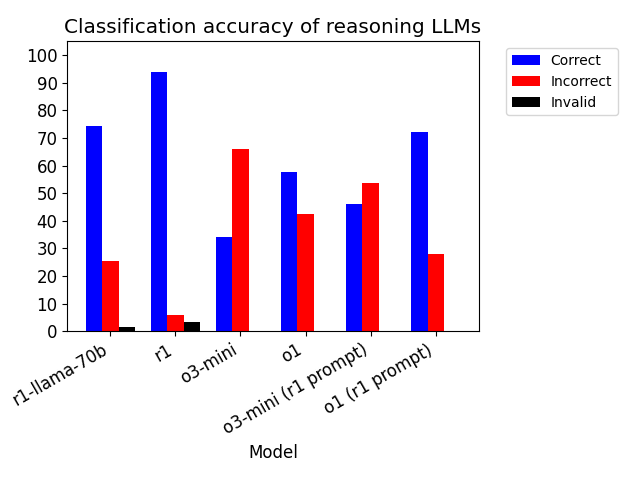}
    \caption{Python builtin identifier swap results for LRMs.}
    \label{FIG:APPENDIX:ADD_IDSWAP_REASONING}
\end{figure}

We evaluate the OpenAI LRMs {\tt o1-2024-12-17} and {\tt o3-mini-2025-01-31} and DeepSeek LRMs {\tt r1} and its distilled version based on Meta {\tt Llama-3.3-70B-Instruct} on the Python builtin identifier swap benchmark. Due to the high cost and low speed of inference, for all these models except {\tt o3-mini-2025-01-31} we only evaluate $10\%$ of the test set.
For the OpenAI models we evaluate both on the default prompt and the CoT-style prompt suggested for DeepSeek-r1.
We report the results in Figure \ref{FIG:APPENDIX:ADD_IDSWAP_REASONING}.

The LRMs are much stronger than {\tt gpt-4o-mini}, {\tt gpt-4.1-nano} and our approach, with the full DeepSeek-r1 reaching 94.0\% accuracy, which is expected given their training and inference costs.

Given sufficient resources, it would be beneficial as a future experiment to use one of these models as the base model for our approach.
We expect that our approach would be complementary to the synthetic data generation techniques used to train these models, resulting in further improvements.

\end{document}